\documentclass[letterpaper, 10 pt, journal, twoside]{IEEEtran}

\usepackage{amsmath,amsfonts}
\usepackage{algorithmic}
\usepackage{algorithm}
\usepackage{array}
\usepackage[caption=false,font=normalsize,labelfont=sf,textfont=sf]{subfig}
\usepackage{textcomp}
\usepackage{stfloats}
\usepackage{url}
\usepackage{verbatim}
\usepackage{graphicx}
\usepackage{cite}
\usepackage{float} 
\usepackage[justification=centering]{caption}
\usepackage{amsmath}
\usepackage{color}
\usepackage{xcolor, soul}
\sethlcolor{yellow}
\definecolor{babyblueeyes}{rgb}{0.63, 0.79, 0.95}
\usepackage{hyperref}
\usepackage{titlesec}
\usepackage{graphics}
\usepackage{ulem}
\usepackage{cancel}

\title{\LARGE \bf
Multi-Modal \\ Multi-Task (3MT) Road Segmentation}

\vspace{-7mm}
\author{Erkan Milli$^{1}$, Özgür Erkent$^{2}$, Asım Egemen Yılmaz$^{1}$
\thanks{\footnotesize{
This work was supported by TUBITAK under EU Commission Horizon 2020 Marie Skłodowska-Curie Actions Cofund program Circulation2 Scheme. The numerical calculations reported in this paper were partially performed at TUBITAK ULAKBIM, (TRUBA).}}
\thanks{\footnotesize{$^{1}$ Erkan Milli and Asım Egemen Yılmaz are with the Department of Electrical and Electronics Engineering, Ankara University, 0600 Ankara, Turkey (e-mail: emilli@ankara.edu.tr; asimegemenyilmaz@yahoo.com).}}
\thanks{\footnotesize{$^{2}$ Özgür Erkent is with Computer Science Department, Hacettepe University, 0600 Ankara, Turkey (e-mail: ozgurerkent@hacettepe.edu.tr).}}
}

\begin{document}

\maketitle

\vspace{-4mm}
\begin{figure*}[b]
    \centering
    \captionsetup{justification=justified}
    \includegraphics[width=15.1cm]{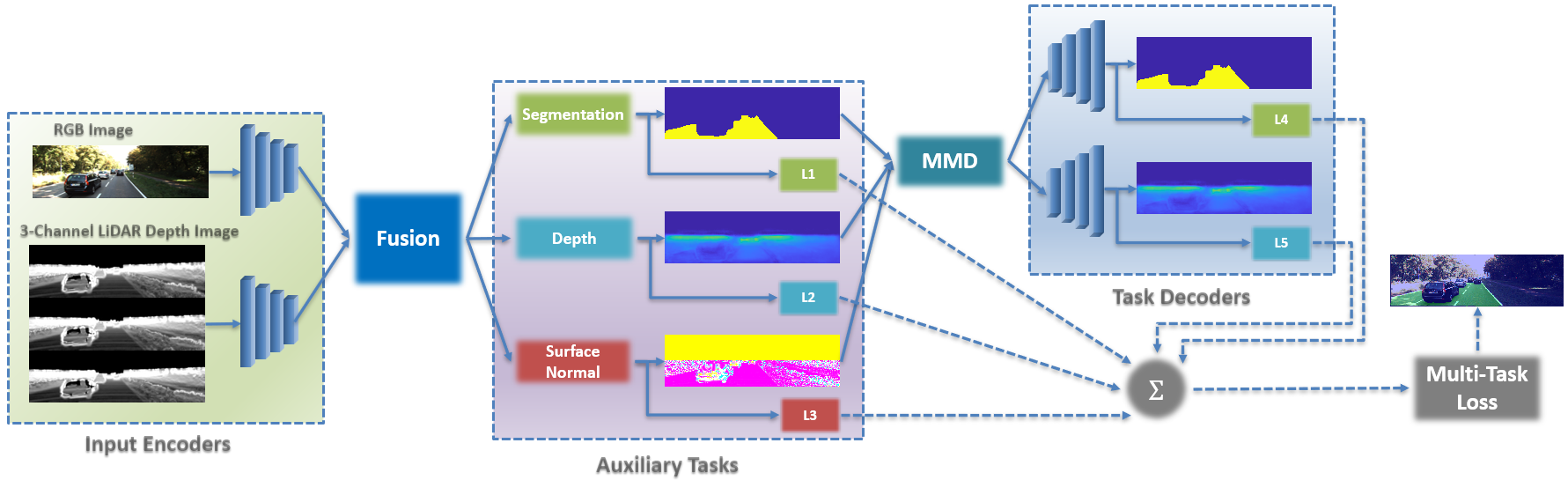}
    \caption{Inputs are RGB images and 3-Channel LiDAR depth images. The shared features coming out of the fully convolutional encoders are passed through the fusion block and sent to the task-specific heads. Initial task estimations in different scales are performed in task-specific heads and the information obtained from this is sent to the multi modal distillation (MMD) block. In MMD, it is aimed to improve the final predictions in the target tasks' decoders by increasing the interaction between the tasks. L1, L2, L3 represent the losses of auxiliary tasks, while L4 and L5 represent the losses of the target tasks, and the Multi-task loss is equal to the sum of all these losses.}
    \label{fig:Total-Model}
\end{figure*}

\begin{abstract}
Multi-modal systems have the capacity of producing more reliable results than systems with a single modality in road detection due to perceiving different aspects of the scene.
We focus on using raw sensor inputs instead of, as it is typically done in many SOTA works, leveraging architectures that require high pre-processing costs such as surface normals or dense depth predictions. By using raw sensor inputs, we aim to utilize a low-cost model that minimizes both the pre-processing and model computation costs.
This study presents a cost-effective and highly accurate solution for road segmentation by integrating data from multiple sensors within a multi-task learning architecture. A fusion architecture is proposed in which  RGB and LiDAR depth images constitute the inputs of the network. Another contribution of this study is to use IMU/GNSS (inertial measurement unit/global navigation satellite system) inertial navigation system whose data is collected synchronously and calibrated with a LiDAR-camera to compute aggregated dense LiDAR depth images. It has been demonstrated by experiments on the KITTI dataset that the proposed method offers fast and high-performance solutions. 
We have also shown the performance of our method on Cityscapes where raw LiDAR data is not available. The segmentation results obtained for both full and half resolution images are competitive with existing methods. Therefore, we conclude that our method is not dependent only on raw LiDAR data; rather, it can be used with different sensor modalities.
The inference times obtained in all experiments are very promising for real-time experiments. The source code is publicly available at \url{https://github.com/ErkanMilli/3MT-RoadSeg}.
\end{abstract}
\section{Introduction}
During autonomous driving, vehicles must first perceive their environment in order to make reliable decisions. A combination of some sensors such as camera, LiDAR, Radar, ultrasonic sensors, etc., can be used for environmental sensing. Different sensors that are used to perceive the environment can simultaneously detect different physical properties of the environment. The information from these multi-modal sensors is of great advantage for consistent and reliable detection of the environment during autonomous driving.
However, the data acquired by the different sensors could be dissimilar in terms of temporal and spatial resolution, data format, and geometric alignment. When different sensors are desired to be used together, pre-pocess may be necessary to eliminate these differences.

Many different methods from past to present for road segmentation have been suggested. \cite{ref5} proposed to work with the Fully Convolutional Network (FCN) for the first time to solve the semantic road segmentation problem, and this study has been a guide for further researches. Studies after this method can be classified as methods that use only RGB \cite{ref6}, \cite{ref7}, \cite{ref14}, \cite{ref15} or only LiDAR \cite{ref16}, \cite{ref17} for the road segmentation task. However, when only camera or only LiDAR is used, some disadvantages arise. When the camera is used alone, the system works better in daylight; reflections and shadows on the road may create problems for segmentation; while, the usage of the sparse point cloud alone, especially in the open area or under rainy or snowy weather conditions, reduce the segmentation performance. In order to eliminate these disadvantages, studies were carried out in which RGB and LiDAR are used together \cite{ref4, ref18, ref19, ref20} by sensor fusion. Road segmentation is still not yet fully solved due to difficulties in different settings as mentioned above. 

Since the last decade, multi-task learning (MTL) methods have been used frequently in detection and classification, detection and segmentation, segmentation and depth estimation task pairs \cite{ref13}. Although, the use of MTL in road segmentation studies is still limited, some studies have shown that it can improve the performance under challenging conditions. For example, \cite{ref23} use bird’s-eye-view (BEV) scene layout estimation, depth estimation and pose estimation with visual odometry as tasks in their MTL model. MultiNet \cite{ref31} includes classification, detection and semantic segmentation tasks. Qian \textit{et al.} \cite{ref33} simultaneously detects drivable areas, lane lines, and traffic objects in the proposed MTL architecture. In another framework \cite{ref21}, the tasks of road segmentation, lane line segmentation, and scene classification are addressed simultaneously. \cite{ref22} uses a MTL model that includes three different perception tasks: traffic object detection, drivable road area segmentation, and lane detection. In recent road segmentation architectures \cite{ref8, ref9, ref10}, which do not have MTL structure, in addition to RGB inputs, disparity/depth data densified by preprocessing is also used as input. Thus, it is aimed to increase the semantic segmentation performance by supporting the visual features with additional features. However, while this increases accuracy, model size, inference time and pre-processing time also increase. It is seen that these disadvantages are eliminated by using models with MTL architecture.
The combination of all these observations has led to the development of the proposed architecture for addressing the road segmentation problem. When segmentation is performed using only a single modal RGB input, it is evident that the performance is highly dependent on the lighting conditions in the scene. This situation prompts us to introduce a second input to the system: LiDAR data. The system performs better with these two inputs; however, it is observed that when there are obstacles such as walls and houses adjacent to the road, these obstacles cannot be distinguished from the road, resulting in false negatives. Therefore, it could be beneficial to develop an architecture that can predict both depth and surface normals (SNs) while segmenting the road: MTL architecture. In the final stage, our architecture utilizes multi-sensor data to improve resilience to poor lighting conditions in the scene and enhances the performance by estimating depth and SNs. Through the method we have developed, all these operations can be accomplished with minimal computational costs. Our approach is simple, and our configuration is set to improve semantic segmentation performance.

In this paper, our objective is to enhance the performance of road segmentation by employing a fusion of RGB and LiDAR data, leveraging multiple sensor types, and exploring the interaction between visual tasks within the MTL framework. Fig. \ref{fig:Total-Model} shows the overview of our proposed method for road segmentation. The model consists of three parts: 
i) Two backbones have two separate inputs, RGB images and 3-channel LiDAR depth images. 
ii) Features from the two backbones are combined and passed through the fusion function and become input to MTI-Net \cite{ref13}, which is used as an MTL model. 
iii) MTI-Net \cite{ref13} consists of three levels in itself. Initial predictions are made for different scales in task-specific heads with features extracted from the backbone in the first stage. The middle-level features that come out of here are passed through the multi-modal distillation block before final predictions are made, so that information between tasks is shared. In the last part, the final predictions are performed.

Our contributions can be listed as follows:
\begin{itemize}
    \item Usage of 3-channel depth images that correspond to current point cloud and the two transformed point clouds from the previous time steps $t-1$ and $t-2$. They are obtained by using transformations computed from IMU/GNSS inertial navigation system measurements via Bayesian filters. This transformation is expected to highlight the differences between the dynamic regions and the static ones. 
    Using a two-stage MTL network for road segmentation with raw sensory data result in a faster run-time. We also show that it can be used for different kinds of sensory data in the experiments part.
    \item As a minor contribution, we also show a new approach for sensory fusion for an MTL network.
\end{itemize}

\titlespacing\subsection{0pt}{2pt plus 2pt minus 2pt}{2pt plus 2pt minus 2pt}\vspace{-2mm}
\section{RELATED WORK}
\subsection{Road Segmentation}
Road segmentation is an active research problem. SOTA (state-of-the-art) results \cite{ref4, ref8, ref9, ref10} have been provided by recent methods.
PLARD \cite{ref4} proposes a non-multi-task but multi-modal architecture to improve road segmentation performance. It is powerful method in terms of segmentation results and offers a robust solution by combining visual features with LiDAR features but has a large model-size and for this reason it requries powerful hardware. This is due to the nature of the network architecture used in this method. The multi-layered fusion architecture of PLARD \cite{ref4} intertwined with the backbone network and the use of the feature space adaptation module, which adapts LiDAR features to visual features, increases the computational cost of the network. In this work, despite utilizing a denser point cloud compared to PLARD \cite{ref4}, we introduce a single-layer fusion structure after the backbone network and a network architecture that entails lower computational expenses in contrast to PLARD \cite{ref4}.

SNE-RoadSeg \cite{ref9} focuses on improving road segmentation accuracy by incorporating SN information into semantic segmentation. Although it achieves high accuracy, its model inference time is slightly high. SNE-RoadSeg+ \cite{ref10} is a refined version of SNE-RoadSeg \cite{ref9} that reduces the model inference time. USNet \cite{ref8} utilizes a network architecture with a symmetric structure and leverages uncertainty information with high accuracy and low model inference time.  SNE-RoadSeg \cite{ref9}, SNE-RoadSeg+ \cite{ref10}, and USNet \cite{ref8} all take RGB and pre-processed depth as inputs. These three methods share a common disadvantage, which is the need for depth input pre-processing and the resulting high-cost preparation time. Therefore, the applicability of these methods for real-time operations is not straightforward. This is because, in real-time scenarios, they would need to predict depth during execution, which would significantly increase the total runtime. In our proposed method, we utilize images obtained from LiDAR point clouds as inputs. The preparation time for each image is significantly low, resulting in a real-time execution. Detailed information is provided in the experiments.

\subsection{Multi Task Learning}
The basic idea in MTL is that different learning tasks with shared representations could be related. MTL proposes that using these representations can improve learning efficiency and prediction accuracy over single-task learning \cite{ref12}. MTL can be considered as a form of inductive transfer, and inductive transfer can also help develop a model by choosing among multiple hypotheses with the help of an inductive bias to be defined \cite{ref26}. During MTL, inductive bias is provided by auxiliary tasks, causing the model to favor hypotheses that explain more than one task \cite{ref26}. The next important thing is how to associate and select the auxiliary tasks with the main task. We can associate visual tasks in the same scene with each other. These tasks can be complementary and regulatory for each other. SNs and depth estimation are good examples of complementary tasks because the information we want to infer can be derived directly from each other \cite{ref13}. In this study, while trying to find drivable areas with semantic segmentation, depth is used as another task in the MTL model. SNs, help to increase semantic segmentation performance as an auxiliary task. The loss for any MTL model can be calculated by taking the arithmetic average of the losses calculated separately for different tasks.

\vspace{-2mm}
\section{METHODOLOGY}
\subsection{Problem Definition}
We can consider the road segmentation task as assigning a road/non-road label to each pixel in a perspective camera image. When LiDAR data $L$ and RGB images $I$ data are used as inputs, the road detection problem is solved by optimizing the following objective \cite{ref4}: 
\begin{equation}
\label{deqn_ex1a}
\underset {W}{\text{arg min}} \sum_{i} \mathcal{L}(f(I_{i},L_{i};W),Y)
\end{equation}
\noindent where $i$ represents training samples, $Y$ is the ground-truth label, $f$ is road segmentation function, $W$ is the model parameters and $\mathcal{L}$ defines loss function. The latter, which will now be used for the multi-task learning model, is a combined loss containing all single-task losses instead of being defined for a single task. For $T$ different single tasks, where $X$ is input and $Y_{T}$ is task-specific labels, the combined loss function can be defined as:
\begin{equation}
\label{deqn_ex1a}
 \mathcal{L}_{total}(X, Y_{T};W_{T}) = \sum_{i=1}^{T}	\lambda_{i} \mathcal{L}_{i} (X,Y_{i}; W_{i})
\end{equation}
\noindent where  $\mathcal{L}_{i}$ is different single-task losses, and $\lambda_{i}$ is task-specific weightings. Three different tasks are performed by the MTL model: semantic segmentation, depth estimation and SN estimation. It is been reported that it is more difficult to minimize the multi-task loss, which includes the losses of different tasks, than to converge a single task loss \cite{ref34}. For a limited number of tasks, the weighting of task losses can be optimized by  trial-and-error \cite{ref35}. The loss weights are regularized to bring them approximately to the same scale. The task weights are tuned on the training set, to maximize the MaxF (maximum F1-measure) score calculated for road segmentation. 
In multi-task learning architecture, it is desirable to have a common representation in the early parts of the network, while in later parts of the network, tasks are solved in heads specific to each task. This is most commonly accomplished as an encoder-decoder construct where each task represents a decoder specific to the representation provided by the common encoder \cite{ref27}. 

\subsection{Fusion Model}
In this paper, an architecture consisting of a single data fusion layer is proposed. First, 3-channel LiDAR depth images are computed using IMU/GNSS navigation 
system poses, and RGB images are fed into the backbone network. Then the features calculated for both inputs are passed through the fusion block:
\begin{equation}
F_{fuse} =  F_{rgb} + \alpha F_{lidar}
\end{equation}
\noindent $F_{lidar}$ is LiDAR features and $F_{rgb}$ is RGB features, $F_{fuse}$ is fusion features, and $\alpha$ is LiDAR weighting coefficient in fusion function and is used as design parameter.

\subsection{LiDAR Registration-Aggregation}
Before the fusion process performed in the feature space, we align the LiDAR and RGB images in the data space by projecting the 3D LiDAR points onto the 2D image plane. All of the operations carried out at this stage can be expressed as LiDAR data registration. The transformation matrix used to project the LiDAR point cloud onto the 2D image plane by:
\begin{equation}
T_{lidar}^{cam} = \begin{bmatrix}  R_{lidar}^{cam} &  t_{lidar}^{cam} \\ 
0_{3}  &1\end{bmatrix}
\end{equation}
\noindent where $T_{lidar}^{cam}$ represents the transformation from LiDAR to camera, $R_{lidar}^{cam}$ is the rotation between LiDAR and camera and $t_{lidar}^{cam}$ is the translation between LiDAR and camera. By projecting a point $x$ from the 3D LiDAR point cloud onto the 2D image plane, the y-point is calculated as:
\begin{equation}
y =  P_{Rect} R_{Rect} T_{lidar}^{cam} x
\end{equation}
\noindent where $P_{Rect}$ denotes projection matrix after rectification, $R_{Rect}$ is the camera rectification matrix.

The ADI (altitude difference image) method in PLARD \cite{ref4} is used to generate the network inputs, which are referred to as LiDAR depth images. During the calculation of the ADI, the Z channel of the LiDAR is considered the altitude. Altitude changes based on offsets between two different positions are included in the process. Thus, ADIs are obtained using altitude differences, and road features are preserved in LiDAR data, making it easier to find drivable areas \cite{ref4}. Altitude difference-based transformation calculates the $V_{xy}$ pixel value of a point  (x,y):
\begin{equation}
V_{x,y} =   \frac{1}{M} \sum_{N_x,N_y} \frac{|Z_{x,y} - Z_{N_x,N_y}|}{\sqrt{(N_x-x)^2 + (N_y-y)^2}}
\end{equation}
\noindent where $Z(x,y)$ represents the elevation of the LiDAR data point projected onto $(x,y)$, $(N_x,N_y)$ indicate locations in the neighbourhood of $(x,y)$, and $M$ denotes the overall count of neighboring positions. This can be considered as calculating the average absolute values of the altitude gradients of the points projected onto the 2D image plane.  Calculations are made in a certain neighborhood window and a neighboring pixel can be easily eliminated if it is uncorrelated in 3D \cite{ref4}.

Point cloud's sparsity degrades the performance of tasks such as semantic segmentation and surface reconstruction. We overcome this problem by using a point cloud accumulation strategy \cite{ref30}. With the method we developed, improvements to the ADI method that will increase the density of LiDAR data are suggested. In order to use this method, data collected from an IMU/GNSS whose data were collected synchronously with the camera and LiDAR and calibrated according to these sensors is needed. Since the KITTI dataset contains IMU data but the Cityscapes dataset does not \cite{ref3}, we had the chance to apply this innovation only to experiments for KITTI and compare the differences in between these two datasets. 

\begin{figure}[h]
    \centering
    \includegraphics[width=7.5cm]{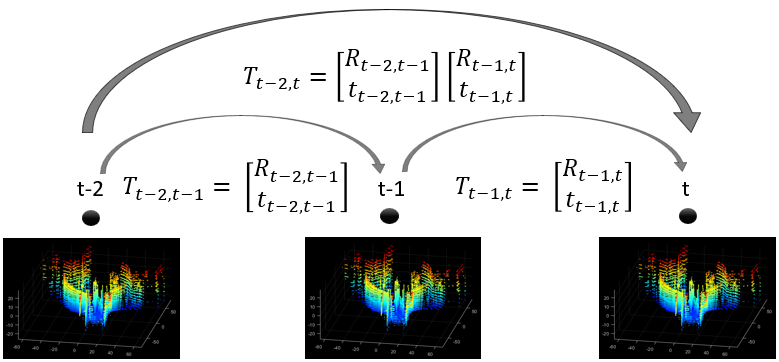}
    \caption{Computation of 3-Channel ADI}
    \label{fig:3-Chn ADI}
\end{figure}

Fig. \ref{fig:3-Chn ADI} denotes the calculation of the densified depth images. The performed operations are as follows: A single depth image is obtained for each of these two time steps by applying transformations that will transform the point clouds at $(t-2)$ and $(t-1)$ into point clouds at $(t)$. For the point cloud at time $(t)$, the depth image of that moment is computed without any transformation. Thus, these depth images computed for three time steps are combined to obtain a single 3-channel image for training moment $t$ and fed to the network. 

\vspace{-4mm}

R expressed in Fig. \ref{fig:3-Chn ADI} is the rotation matrix and T is the translation matrix. It is seen that the accuracy of R and T found with the IMU/GNSS navigation system pose shared in KITTI is not sufficient for the developed method. Therefore, R and T are found by applying Kalman integration and a more precise navigation algorithm to IMU/GNSS data shared in KITTI. R matrix is found from estimated euler attitudes, T matrix is found from estimated 3D positions. Details about the proposed method and the equations used during implementation can be found in the Appendix.

\vspace{-2mm}
\begin{figure}[h]
    \centering
    \includegraphics[width=\linewidth]{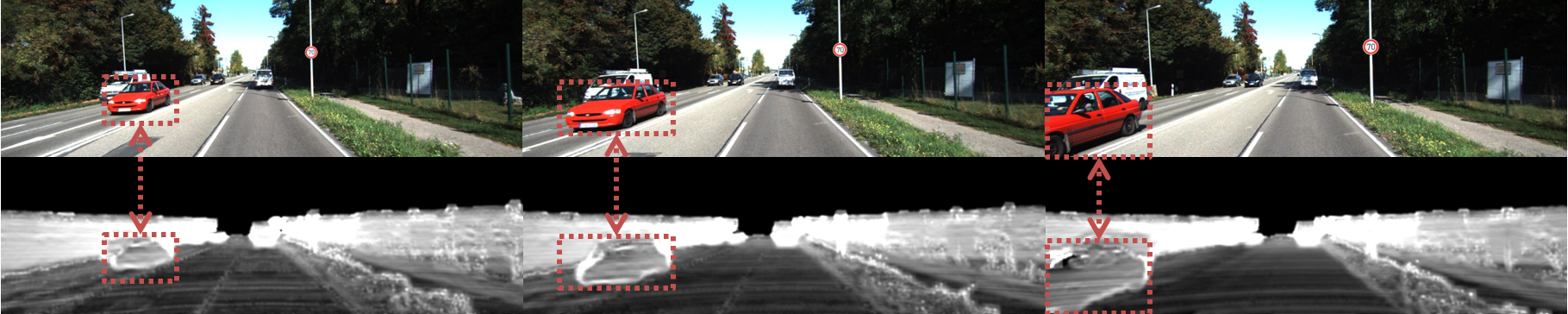}
    \caption{3-channel ADI time stream (moments $t-2$, $t-1$ and $t$)}
    \label{fig:Proposed ADI}
\end{figure}



Fig. \ref{fig:Proposed ADI} shows the RGB images in three consecutive time steps and their corresponding depth images. These consecutive depth images combine to form the 3-channel ADIs we proposed. While the depth images at $(t-2)$ and $(t-1)$ moments are obtained from the transformations of the respective moments at time $(t)$, no transformation is applied to the image at time $(t)$.  

\begin{figure}[h]
    \centering
    \includegraphics[width=\linewidth]{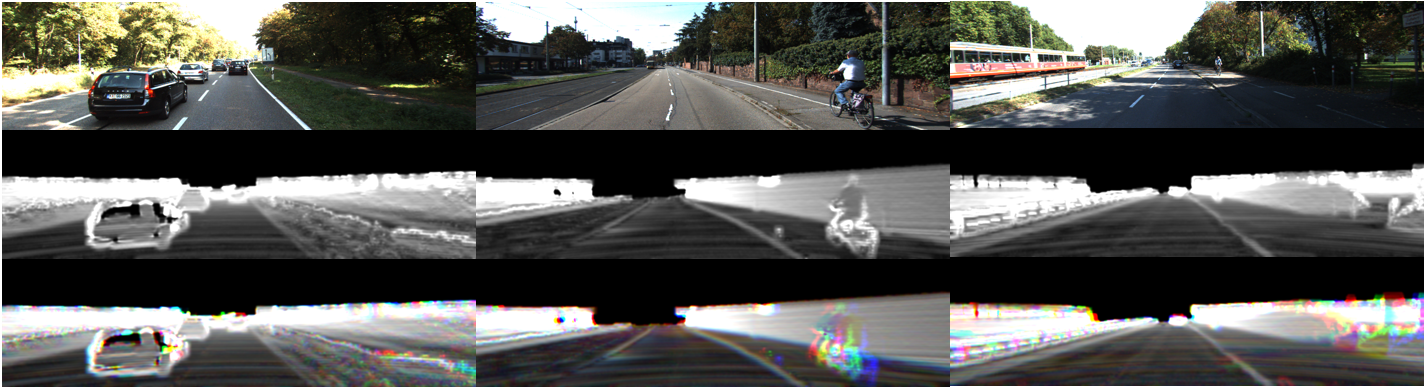}
    \caption{Top to buttom: RGB images, ADIs in PLARD \cite{ref4} and 3-channel ADIs proposed by us}
    \label{fig:ADI comparison}
\end{figure}

\vspace{4mm}
Fig. \ref{fig:ADI comparison} shows a comparison of the ADIs in PLARD \cite{ref4} with the 3-channel ADIs we propose. It is evident that the point cloud densities of 3-channel ADIs are increased compared to standard ADIs. Thus, it provides performance increase for road segmentation by densifying the point cloud. In addition, while 3-channel ADIs have colorations, standard ADIs do not. Colorations represent the differences in the transformations of the point clouds of the moving regions from the previous two timesteps into time (t). They are much greater when INS data is used directly instead of using our proposed method. See Appendix for details.

\subsection{Model Architecture}
An architecture is proposed in which multi-sensor data is fused and passed through the MTL network. Accordingly, RGB images and 3-channel ADIs densified with IMU/GNSS poses are separately passed through the backbone network HRNet \cite{ref25}. Specifically, HRNet \cite{ref25} has SOTA accuracy, it can process attributes of different scales simultaneously, and is computationally efficient despite its high accuracy. The features obtained at the backbone output are fed into the MTL network by passing through a feature-based fusion function. Primitive features passed through the feature-based fusion function are sent to task-specific heads for initial task estimations.

\begin{figure}[h]
    \centering
    \captionsetup{justification=justified}
    \includegraphics[width=6.9cm]{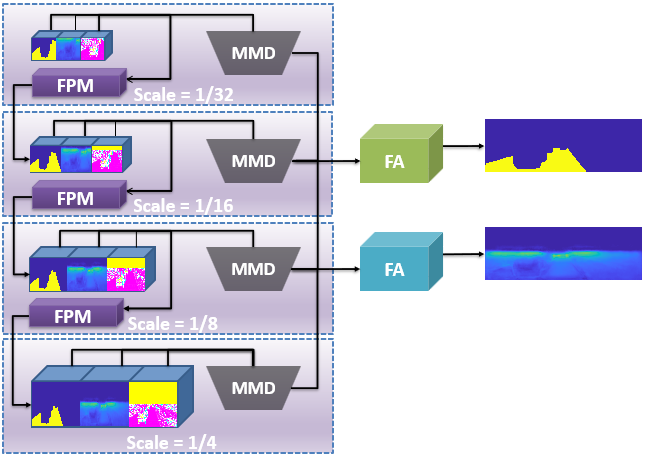}
    \caption{Overview of modules used in MTL architecture}
    \label{fig:MTI-Net}
\end{figure}

In this study, a customized version of MTI-Net proposed by Vandenhende \textit{et al.} \cite{ref13} is used by us to apply to the road detection problem at KITTI. While feature aggregation and distillation blocks are used as in MTI-Net \cite{ref13}, 3-channel ADIs and RGBs before these blocks are passed through a fusion process implemented in the feature space and a multi-modal and temporal structure is obtained. Thus, the proposed multi-modal and multi-task road segmentation (3MT-RoadSeg) structure emerges. Since there is no LiDAR data in the Cityscapes dataset, MTI-Net \cite{ref13} is used without modifications in some experiments for this dataset. Fig. \ref{fig:MTI-Net} denotes the  MTI-Net \cite{ref13} architecture.  MTI-Net \cite{ref13} is basically based on the prediction-and-distillation network PAD-Net \cite{ref24}. PAD-Net \cite{ref24} is an MTL model developed to perform depth estimation and scene parsing from a single RGB image with the help of monocular depth prediction and SN estimation auxiliary tasks. The main difference between  MTI-Net \cite{ref13} and PAD-Net \cite{ref24} is the distillation modules they use. The PAD-Net \cite{ref24} architecture effectively uses supplementary information from the intermediate estimates of tasks associated with a MMD. In MTI-Net \cite{ref13}, task features are distilled separately at each scale, with the idea that tasks may affect each other differently for different sizes of receptive fields. After distillation, features from all scales are collected to make final estimates. Feature propagation module (FPM) uses task-specific and scale-specific attention mechanisms to selectively propagate features across different tasks and scales. FPM is used to transmit distilled information from lower resolution task features to higher ones. This allows the network to learn and solve multiple tasks simultaneously by leveraging shared information across different tasks and scales. Feature aggregation (FA) combines and integrates features from different scales which enhances the network's representation learning capabilities. This results in learning more robust and discriminative representations which improves the performance of the network.



\begin{table*}[b]
\centering
\caption{Road Segmentation Results for KITTI Validation Set\\
"*": result with standard ADIs}\label{table-1}
\resizebox{14.5cm}{!}{%
\begin{tabular}{ c c c c c c c c c c c }
 \hline
 \textbf{Input}         & \textbf{Task} & \textbf{Aux. Task} & \textbf{MaxF(\%)} & \textbf{AP(\%)} & \textbf{PRE(\%)} & \textbf{REC(\%)} & \textbf{FPR(\%)} & \textbf{FNR(\%)} & \textbf{Runtime(s)} \\\hline\hline
 rgb                    & S     & S     & 96.92          & 92.38          & 96.54          & 97.30          & 0.77          & 2.70             & \textbf{0.044}   \\
 lidar                  & S     & S     & 95.99          & 91.93          & 95.59          & 96.40          & 0.98          & 3.60             & \textbf{0.044}   \\
rgb + lidar            & S     & S     & 97.24          & 92.30           & 97.12 & 97.37  & 0.64           & 2.63           & 0.072   \\
 rgb                    & S+D   & S+D   & 96.99          & 92.40          & 96.70          & 97.27          & 0.73          & 2.73             & \underline{0.054}   \\
 lidar                  & S+D   & S+D   & 96.21          & 92.37          & 96.07          & 96.36          & 0.87          & 3.64             & \underline{0.054}   \\
 rgb + lidar            & S+D   & S+D   & \underline{97.31} & 92.44       & 97.15 & \underline{97.48}       & 0.63          & \underline{2.52} & 0.084   \\
 rgb                    & S+D   & S+D+N & 97.09          & \textbf{92.51} & 97.11          & 97.06          & 0.64          & 2.94             & 0.059   \\
 lidar                  & S+D   & S+D+N & 96.40          & 92.32          & 96.16          & 96.64          & 0.85          & 3.36             & 0.059   \\
 rgb + lidar            & S+D   & S+D+N & \textbf{97.39} & 92.45 & \underline{97.27} & \textbf{97.51}       & \underline{0.61}  & \textbf{2.49} & 0.089   \\
 rgb + lidar*           & S+D   & S+D+N & 97.28          & \underline{92.47} & \textbf{97.28} & 97.27       & \textbf{0.60} & 2.73             & 0.074   \\\hline
\end{tabular}}
\end{table*}

\vspace{-4mm}
\section{EXPERIMENTS}
\titlespacing\subsection{0pt}{6pt plus 2pt minus 2pt}{6pt plus 2pt minus 2pt}
KITTI and Cityscapes datasets were used in the experiments. 289 training and 290 test images exist in the KITTI road-data. These images are categorized as UM (urban marked roads), UMM (urban multiple marked roads), and UU (urban unmarked roads). Since  only labels corresponding to the training set in the KITTI is available, we use \%70 of the training set as training; and the remaining \%30  as a validation set. In the experiments performed with the whole training set, the road masks found to measure the test results were converted to BEV and sent to the submission page of KITTI \cite{ref2}. Cityscapes has labels for 19 different classes and contains dynamic objects, changing scene layouts and backgrounds. 2975 images in the data belong to the training set, 500 images to the validation set \cite{ref3}. 

\subsection{Evaluation Metrics}
We utilize the commonly employed pixel-level segmentation metrics for road segmentation to conduct quantitative assessment. These are MaxF, AP (average precision), PRE (precision rate), REC (recall rate), FPR (false positive rate) and FNR (false negative rate). These metrics are shared in detail with \cite{ref1}. When evaluating depth prediction results, we employ the widely used quantitative evaluation metric SILog (scale invariant logarithmic error) \cite{ref38}.

\vspace{-1mm}
\begin{table}[h]
\centering
\caption{Depth Results for the KITTI Road Set}\label{table-2}
\resizebox{8cm}{!}{%
\begin{tabular}{ c c c c c c c c c c c }
 \hline
 \textbf{Methods}          & \textbf{Task} & \textbf{Aux. Task}      & \textbf{SILog}     \\\hline\hline
 MTI-Net  \cite{ref13}     &    S+D        &      S+D+N              &     15.34         \\\hline
 3MT-RoadSeg (rgb + lidar) &    S+D        &      S+D                &     \textbf{10.42}            \\\hline
\end{tabular}}
\end{table}

\vspace{-2mm}
\subsection{Implementation Details}
During the training, a single NVIDIA Tesla V100 GPU with 16 GB memory is used and the network is implemented with the PyTorch framework. In our experiments with the  MTI-Net \cite{ref13} model, the HRNet-32 \cite{ref25} backbone is used for both KITTI and Cityscapes datasets. The loss function is optimized using the Adam optimizer with a learning rate of 1e-4.

\begin{table*}[t]
\centering
\caption{Comparison of Different Methods in KITTI Benchmark\\
"*": result with standard ADIs, "\textsuperscript{+}": denotes Multi-Scale version}\label{table-3}
\resizebox{15cm}{!}{%
\begin{tabular}{ c c c c c c c c c c c }
 \hline
 \textbf{Methods}                 &   \textbf{Input}     & \textbf{Multi-tasks} & \textbf{MaxF(\%)} & \textbf{PRE(\%)} & \textbf{REC(\%)} & \textbf{FPR(\%)} & \textbf{FNR(\%)} & \textbf{Runtime(s)} \\\hline\hline
 s-FCN-loc \cite{ref7}   & rgb         &  -      & 93.26          & 94.16          & 92.39          & 3.16             & 7.61             & 0.40            \\
 MultiNet  \cite{ref31}  & rgb         &  C+OD+S & 94.88          & 94.84          & 94.91          & 2.85             & 5.09             & 0.17            \\
 RBNet  \cite{ref6}     & rgb         &  -      & 94.97          & 94.94          & 95.01          & 2.79             & 4.99             & 0.18            \\
 RBANet  \cite{ref15}    & rgb         &  -      & 96.30          & 95.14          & \underline{97.50} & 2.75          & \underline{2.50} & 0.16            \\
 LidCamNet \cite{ref18}  & rgb + lidar &  -      & 96.03          & 96.23          & 95.83          & 2.07             & 4.17             & 0.15            \\
 CLCFNet  \cite{ref20}   & rgb + lidar &  -      & 96.38          & 96.38          & 96.39          & 1.99             & 3.61             & \textbf{0.02}   \\
 PLARD  \cite{ref4}      & rgb + lidar &  -      & 96.83          & 96.79          & 96.86          & 1.77             & 3.14             & 0.16            \\
 PLARD\textsuperscript{+} \cite{ref4}      & rgb + lidar &  -      & \underline{97.03} & \underline{97.19} & 96.88    & \underline{1.54} & 3.12             & 1.50            \\
 SNE-RoadSeg\cite{ref9}  & rgb + depth &  -      & 96.75          & 96.90          & 96.61          & 1.70             & 3.39             & 0.18          \\
 DFM-RTFNet \cite{ref32} & rgb + depth &  -      & 96.78          & 96.62          & 96.93          & 1.87             & 3.07             & 0.08           \\
 USNet  \cite{ref8}      & rgb + depth &  -      & 96.89          & 96.51          & 97.27          & 1.94             & 2.73             & \textbf{0.02}     \\
 SNE-RoadSeg+\cite{ref10}& rgb + depth &  -      & \textbf{97.50} & \textbf{97.41} & \textbf{97.58} & \textbf{1.43}    & \textbf{2.42}    & 0.08   \\\hline
 3MT-RoadSeg*            & rgb + lidar&  S+D     & 96.60          & 96.46          & 96.73          & 1.95             & 3.27             & \underline{0.07} \\\hline
 \end{tabular}}
\end{table*}
\vspace{-4mm}

\begin{table*}[t]
\centering
\caption{Road Segmentation Results for Cityscapes Validation Set}\label{table-4}
\resizebox{15.5cm}{!}{%
\begin{tabular}{ c c c c c c c c c c c }
 \hline
 \textbf{Input}         & \textbf{Task} & \textbf{Aux. Task} & \textbf{MaxF(\%)} & \textbf{AP(\%)} & \textbf{PRE(\%)} & \textbf{REC(\%)} & \textbf{FPR(\%)} & \textbf{FNR(\%)} & \textbf{Runtime(s)} \\\hline\hline
 rgb (half-res)         & S     & S     & 97.24          & 91.04          & 96.85          & 97.63          & 1.56          & 2.37          & \textbf{0.04} \\ 
 rgb (half-res)         & S+D   & S+D+N & 97.46          & 91.51          & 97.37          & 97.56          & 1.30          & 2.44          & \underline{0.06} \\ 
 rgb + depth (half-res) & S+N   & S+D+N & 97.49          & 91.65          & 97.52          & 97.45          & 1.22          & 2.55          & 0.09          \\ 
 rgb (full-res)         & S     & S     & 97.44          & \textbf{92.07} & \textbf{97.98} & 96.90          & \textbf{0.98} & 3.10          & 0.15          \\ 
 rgb  (full-res)        & S+D   & S+D+N & \underline{97.60} & 91.24       & 97.07          & \underline{98.12} & 1.45       & \underline{1.88} & 0.23          \\ 
 rgb + depth (full-res) & S+N   & S+D+N & \textbf{97.84} & \underline{91.68} & \underline{97.55} & \textbf{98.13} & \underline{1.21} & \textbf{1.87} & 0.36          \\\hline
\end{tabular}}
\end{table*}

\vspace{-3mm}
\begin{table}[t]
\centering
\caption{Comparative results for the Cityscapes Validation Set}\label{table-5}
\resizebox{8cm}{!}{%
\begin{tabular}{ c c c c c c c c c c c }
 \hline
 \textbf{Methods}               & \textbf{MaxF(\%)}  & \textbf{PRE(\%)} & \textbf{REC(\%)}    \\\hline\hline
 FCN \cite{ref5}                & 94.68          & 93.69          & 95.70                     \\
 s-FCN-loc \cite{ref7}          & 95.36          & 94.63          & 96.11                     \\ 
 SegNet \cite{ref11}            & 95.81          & 94.55          & 97.11                     \\ 
 RBANet \cite{ref15}            & \underline{98.00} & \underline{97.87} & \underline{98.13}   \\ 
 USNet  \cite{ref8}             & \textbf{98.27} & \textbf{98.26} & \textbf{98.28}            \\\hline
 3MT-RoadSeg (rgb)              & 97.60          & 97.07          & 98.12                     \\ 
 3MT-RoadSeg (rgb + depth)      & 97.84          & 97.55          & \underline{98.13}         \\\hline
\end{tabular}}
\end{table}

\vspace{7mm}
\subsection{Evaluation Results}

\subsubsection{KITTI Ablation Studies}
In this section, the test results obtained for different input, task and auxiliary task configurations are compared, and the image resolution  is 384$\times1280$. KITTI validation set is used in the experiments and the test results are shown in Table \ref{table-1}. The highest MaxF score is achieved for \textit{rgb+lidar} input, \textit{S (semantic segmentation) + D (depth)} multi-tasks and \textit{S + D + N (surface normal)} auxiliary-tasks configuration. This shows that road semantic segmentation performance is improved with the use of different tasks and auxiliary tasks that use similar features. It is clearly seen that the performance of the multi-task model are better than the single-task model by increasing the interaction between the tasks for the shared features for different tasks. In addition, it is observed that the MaxF scores obtained for experiments with \textit{rgb+lidar} inputs are higher than those that are only \textit{rgb} or \textit{lidar} as input. This reveals to us the contributions of the multi-modality created by fusion operation. However, \textit{rgb+lidar} input configuration has the longer inference time. In the \textit{rgb+lidar*} configuration unlike the \textit{rgb+lidar} configuration, standard ADIs are used instead of the 3-channel ADIs developed by us. The results show that the use of 3-channel ADIs increases MaxF. This comparison shows the contribution of the developed 3-channel ADIs to the road segmentation results.

The results of the depth estimation conducted simultaneously with the road segmentation for KITTI are presented in Table \ref{table-2}. It is evident that fusing 3-channel ADIs with RGB information enhances the depth prediction performance by supporting the depth features as expected. The achievement of such a result solely using RGB is not feasible. It is important to keep in mind that although the network was mainly trained for road segmentation, depth estimation can also be improved with respect to MTI-Net.

\subsubsection{KITTI Benchmark Results}

This section focuses on the KITTI benchmark results obtained against the BEV forms of the road semantic segmentation results of the test set submitted to the KITTI submission page. Since road mapping tags are not shared for the samples in KITTI's test set, the 3-channel ADI method we developed could not be used in the results submitted to the benchmark. Instead, the standard ADI method in PLARD \cite{ref4} is used when generating LiDAR inputs. Table \ref{table-3} contains a comprehensive comparison of our proposed method and other methods on KITTI benchmark. SNE-RoadSeg+\cite{ref10} has the highest MaxF score and uses \textit{rgb+depth} inputs. PLARD \cite{ref4} has the highest MaxF score among studies using \textit{rgb+lidar} inputs. 3MT-RoadSeg has a higher MaxF than all \textit{rgb} input methods and all \textit{rgb+lidar} except PLARD\cite{ref4}. 3MT-RoadSeg (0.07 s) has the shortest model inference time among all studies except CLCFNet \cite{ref20} (0.02 s) and USNet \cite{ref8} (0.02 s). When considering the total runtimes, which are calculated by adding the model inference times to the input pre-processing times, the time efficiency solution offered by 3MT-RoadSeg emerges even more prominently. The high MaxF-score methods SNE-RoadSeg \cite{ref9}, SNE-RoadSeg+ \cite{ref10} and USNet \cite{ref8} utilize the same depth inputs. However, depth computation time is not provided. If a coarse monocular depth estimation method \cite{ref39} is utilized, the inference time can be approximately 0.01 seconds. However, by employing a fine monocular depth estimation method like VA-DepthNet \cite{ref36}, which ranks among the top in the KITTI benchmark, the inference time can reach approximately 0.34 seconds. Nevertheless, considering the expectation of accurate depth estimations used as model inputs, the fine depth estimation method VA-DepthNet \cite{ref36} is employed to obtain the provided results. When the experimental pre-processing time of 0.34 s is added to the model inference times of SNE-RoadSeg \cite{ref9}, SNE-RoadSeg+ \cite{ref10}, and USNet \cite{ref8}, the total runtimes for these methods become 0.52 s, 0.42 s, and 0.36 s, respectively. In DFM-RTFNet \cite{ref32}, one of the inputs is the disparity; hence, the value of 0.13 s is used, which is reported in SOTA disparity estimation method FADNet++ \cite{ref37}. The total runtime for DFM-RTFNet \cite{ref32} is calculated as 0.21 seconds. CLCFNet \cite{ref20} uses LiDAR image as input, whose preparation is similar to our ADI computation method. Therefore, the preparation time of the LiDAR imagery is expected to be approximately 0.01 s. The total runtime for CLCFNet \cite{ref20} can be calculated as approximately 0.03 s. The total runtime of the 3MT-RoadSeg becomes 0.08 s after adding the ADI preparation time. 
The shortest total runtime after CLCFNet \cite{ref20} is 3MT-RoadSeg. The method we propose is very useful for real-time scenarios in terms of total runtime and high accuracy. For the KITTI benchmark, there are no results for any MTL architecture method other than our method 3MT-RoadSeg and MultiNet \cite{ref31}. 3MT-RoadSeg uses \textit{S + D}, and MultiNet \cite{ref31} uses \textit{C (classification) + OD (object detection) + S} tasks. From the results in Table \ref{table-3}, it appears that the road segmentation performance of 3MT-RoadSeg is better than MultiNet  \cite{ref31} and has shorter inference time.

\subsubsection{Cityscapes Ablation Studies}
The results of the experiments for half \textit{($512\times1024$)} and full resolution ($1024\times2048$) inputs in the Cityscapes validation set are shown in Table \ref{table-4}. The MaxF score of the multi-task model, in which more than one task is used together, is higher than the single task model. In the experiments, it is observed that there is a slight improvement in the multi-mode structure with \textit{rgb+depth} input compared to the single-mode structure with only \textit{rgb} input. The reason for this is that depth inputs fed to the network as inputs are obtained from disparities that are shared in the dataset and are quite noisy, although dense. As expected, it is seen that the test results performed at full resolution are better than those performed at half resolution. In addition, the runtimes of half-resolution experiments are lower than those of full-resolution. Table \ref{table-5} shows the results obtained for the different methods for the Cityscapes validation set. From the results, it is seen that the scores obtained for our proposed method are in the third place after USNet  \cite{ref8} and RBANet \cite{ref15}.

\titlespacing\section{0pt}{3pt plus 2pt minus 2pt}{3pt plus 2pt minus 2pt}\vspace{-1mm}
\section{CONCLUSION}
In this work, a multi-modal and multi-task road segmentation approach is proposed. This method leverages densified LiDAR point cloud to enhance segmentation performance. Through the conducted experiments, it becomes evident that incorporating LiDAR-camera fusion and a multi-task architecture, which capitalizes on the interactions between different tasks, significantly improves the segmentation accuracy. Moreover, our developed method demonstrates lower computational costs compared to most SOTA techniques. When considering the computational expenses of pre-processing, the runtime efficiency of our 3MT-RoadSeg model becomes even more remarkable. These advantages make our method highly suitable for real-time applications, making it a favorable choice over other existing methods.

\appendix
\small
This section will focus on how Kalman integration is applied to INS pose data shared in KITTI in the proposed method to densify the LiDAR point cloud. Normally, INS outputs are obtained as a result of integrating IMU and GNSS data with the help of Kalman filter. In the method we developed, it is proposed to apply a Kalman filter to the integrated solution data shared in KITTI once again.

\begin{figure}[h]
    \centering
    \includegraphics[width=\linewidth]{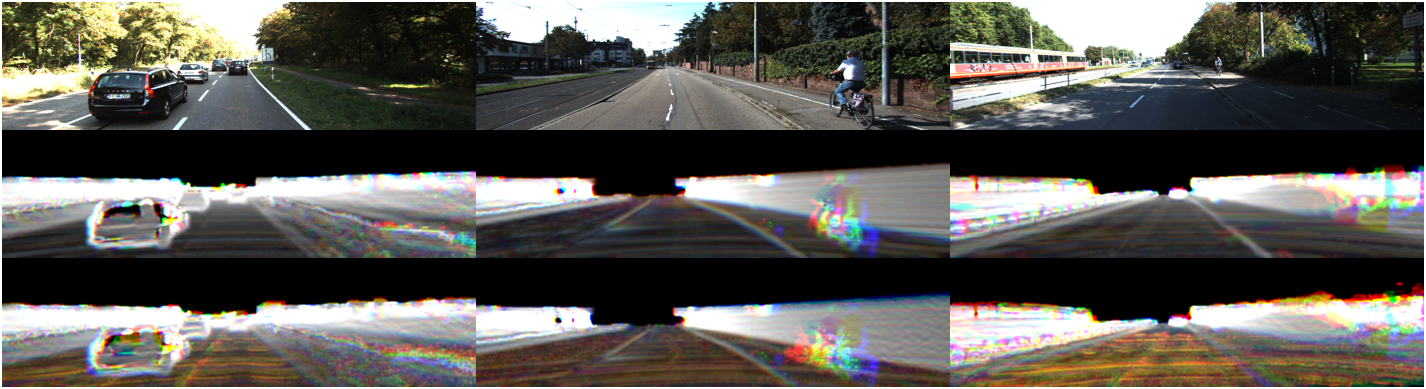}
    \caption{Kalman Filter ADI - OXTS (INS) ADI}
    \label{fig: INS ADI}
\end{figure}

The inertial navigation sensor used when creating the KITTI dataset is the OXTS RT3003. OXTS is a sensor that can provide INS (IMU/GNSS) integrated solution. In the dataset, both IMU outputs from OXTS data and INS outputs integrated with GNSS are shared. When the proposed 3-channel ADIs are generated with OXTS INS position (translation) and INS attitude (rotation) data shared in KITTI, the results are not as desired. The reason for this is thought to be residual errors, especially in position errors. It can now be shown that the source of the errors is using a simple earth model and inertial navigation algorithm during the OXTS IMU/GNSS integration (where the velocity vector $V_{eb}^n$ is a 2D vector instead of a 3D vector, etc.). To eliminate these residual errors, INS data is considered as aiding data and integrated with IMU data. It is seen that this integration process from the results now eliminates the errors and gives the expected results. Fig. \ref{fig: INS ADI} shows the ADIs generated from the OXTS data and the ADIs generated using the integration results proposed by us. Significant color differences are seen in the ADIs calculated from the OXTS data compared to the ADIs calculated in the developed method. This means that there is error in point cloud transformation between consecutive samples. Moires occur in ADIs found directly from OXTS data, especially around moving objects, due to translation errors at moments (t-2) and (t-1). Therefore, the clarity and distinguishability of OXTS ADIs are less than the ADIs found by the developed method. More importantly, the roadside lines in OXTS ADIs are less distinct than in ADIs found by the proposed method. All these advantages show the effectiveness of our proposed method.


\section*{Inertial Navigation Equations}
The inertial navigation equations used in the proposed method are defined in the local navigation frame. The local navigation axes are a Cartesian reference defined with respect to the local horizontal plane that moves with the platform. It is defined according to the ellipsoid, as in the geodetic reference. The X-axis points to the North, the Y-axis to the East, and the Z-axis to the ground. Attitude update equations defined in local navigation frame as \cite{ref29}:

\vspace{-3mm}
\begin{equation}
C_{b}^{n}(+) \approx C_{b}^{n}(-) (I_{3} + \Omega_{ib}^{b}\tau_{i}) - (\Omega_{ie}^{n}(-) + \Omega_{en}^{n}(-)) C_{b}^{n}(-)\tau_{i}
\label{eq:Cbn}
\end{equation}

\noindent where $C_{b}^{n}$ is body-to-navigation-frame coordinate transformation matrix, $\Omega_{ib}^{b}$ is skew-symmetric form of IMU angular-rate vector, and $\tau_{i}$ is time interval. Skew-symmetric matrix of Earth-rotation vector defined as $\Omega_{ie}^n = \omega_{ie}\begin{bmatrix} \cos L_b &  0 &  -\sin L_b \end{bmatrix}\wedge$. The WGS-84 value of the Earth’s angular rate is $w_{ie} = 7.292115\times10^{–\:5}$ rad $s^{-\:1}$ and $'\wedge'$ denotes skew-symmetric transform.

$\Omega_{en}^{n}$ denoted by Eq. \ref{eq:Cbn} is the skew-symmetric matrix of $\omega_{en}^n$ angular rate, $\omega_{en}^n$ calculated as:

\begin{equation}
\omega_{en}^n = \begin{bmatrix}  V_{eb,E}^n/(R_E(L_b)+h_b) \\
-V_{eb,N}^n/(R_N(L_b)+h_b)  \\
-V_{eb,E}^{n}\tan L/(R_E(L_b)+h_b)
\label{eq:w_en_n}
 \end{bmatrix}
\end{equation}

\noindent where $V_{eb}^n$ is the velocity defined on Earth-referenced in local navigation frame axes. $R_N(L_b)$ is the meridian radius of curvature and $R_E(L_b)$ is the normal radius of curvature. They are given by:
\begin{equation}
R_N(L_b) = \frac{R_0(1-e^2)}{(1-e^2\sin^2 L_b)^{3/2}}, 
R_E(L_b) = \frac{R_0}{(1-e^2\sin^2 L_b)}
\end{equation}
The velocity update equation defined on the local navigation frame is as follows \cite{ref29}:
\begin{equation}
\begin{aligned}
v_{eb}^{n}(+) &\approx v_{eb}^{n}(-) + [f_{ib}^{n} + g_{b}^{n}(L_{b}(-), h_{b}(-)) \\
&\quad -(\Omega_{en}^{n}(-) + 2\Omega_{ie}^{n}(-)) v_{eb}^{n}(-)]\tau_{i}
\end{aligned}
\end{equation}
\noindent where $f_{ib}^{n}$ is the specific force measured by IMU ($f_{ib}^{b}$) converted to local navigation frame, $g_{b}^{n}$ is acceleration due to the gravity and calculated using the WGS-84 Earth Ellipsoid model.

Assuming the velocity changes as a linear function of time over the sample time, the position updates are:

\begin{equation}
\begin{aligned}
L_{b}(+) &= L_{b}(-) + \frac{\tau_i}{2}(\frac{v_{eb,N}^{n}(-)}{R_N(L_b(-))+h_b(-)}\\
&\quad+ \frac{v_{eb,N}^{n}(+)}{R_N(L_b(-))+h_b(+))})
\end{aligned}
\end{equation}

\begin{equation}
\begin{aligned}
\lambda_b(+) &=\lambda_{b}(-) + \frac{\tau_{i}}{2}(\frac{v_{eb,E}^{n}(-)}{(R_E(L_b(-)+h_b(-))\cos L_b(-)}\\
&\quad+ \frac{v_{eb,E}^{n}(+)}{(R_E(L_b(+)+h_b(+))\cos L_b(+))})
\end{aligned}
\end{equation}

\begin{equation}
\begin{aligned}
h_{b}(+) = h_{b}(-) - \frac{\tau_i}{2}(v_{eb,D}^{n}(-) + v_{eb,D}^{n}(+))
\end{aligned}
\end{equation}
where, $L_b$ is geodetic latitude, $\lambda_b$ is longitude, and $h_b$ is geodetic height \cite{ref29}.


\begin{figure}[h]
    \centering
    \includegraphics[width=\linewidth]{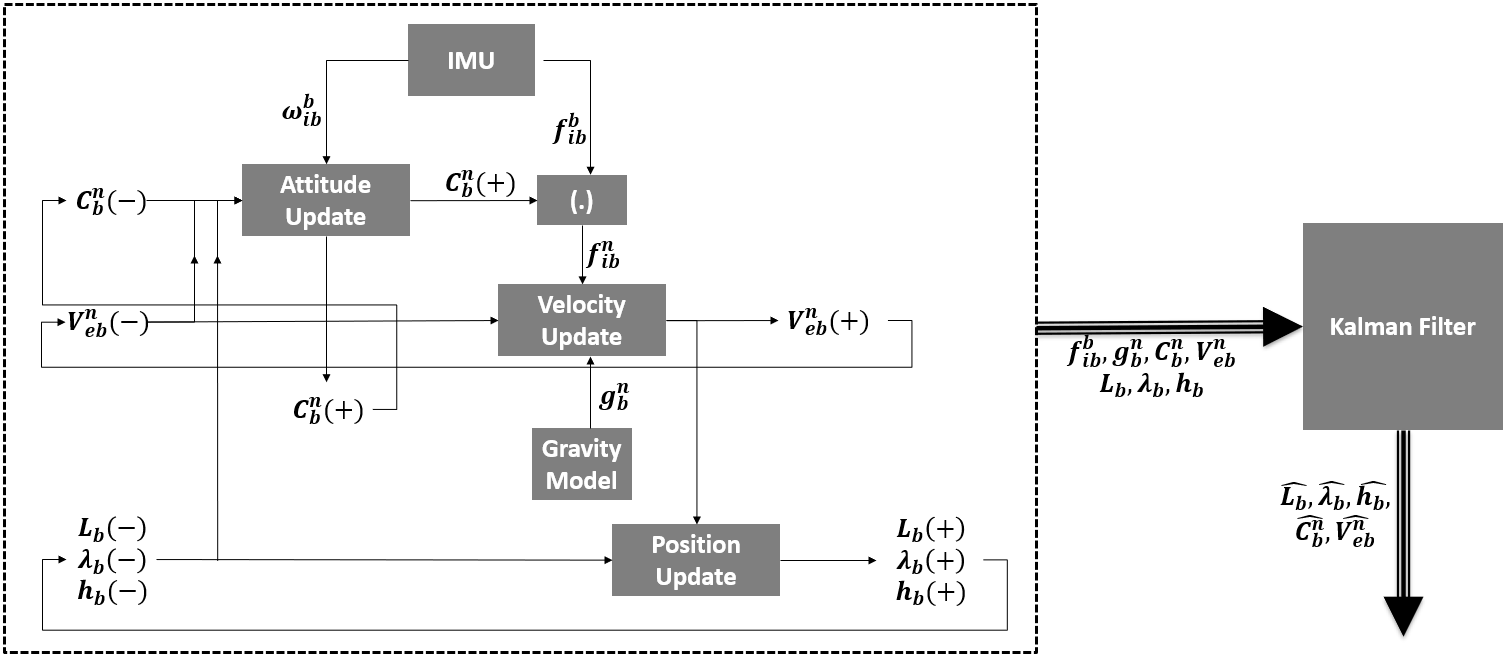}
    \caption{Navigation updates and open-loop Kalman integration model}
    \label{fig:navModel}
\end{figure}
\section*{INS/GNSS Integration Model}

Open loop Loosely Coupled integration model and INS/GNSS Kalman filter algorithm with error state are used during the calculation of the proposed 3-channel ADIs. In the Kalman Filter implementation, the attitude and velocity are Earth-referenced and resolved in the local navigation frame. Position terms are expressed as latitude, longitude and height. Fig. \ref{fig:navModel} shows the block diagram of the GNSS integrated navigation algorithm. The navigation equations (updates) are shown on the left, while the inputs and outputs of the Kalman filter with open loop architecture are shown on the right. The states of the 15-state kalman filter are $\delta\widehat{p}_b$, $\delta \widehat{v}_{eb}^n$, $\delta \widehat{\Psi}_{nb}^n$  respectively, position, velocity  and attitude error states, and $\widehat{b}_a$, $\widehat{b}_g$ are acceleration and gyroscope bias. The state-space equation in continuous time of the system model implemented in the local navigation frame is:

\begin{equation}
 \begin{bmatrix}  \delta\widehat{\dot{p}}_b\\
 \delta \widehat{\dot{v}}_{eb}^n\\ \delta \widehat{\dot{\Psi}}_{nb}^n\\
 \widehat{\dot{b}}_a\\ \widehat{\dot{b}}_g
 \end{bmatrix} = F_{INS}^n \begin{bmatrix}  \delta\widehat{p}_b\\
 \delta \widehat{v}_{eb}^n\\ \delta \widehat{\Psi}_{nb}^n\\
 \widehat{b}_a\\ \widehat{b}_g
 \end{bmatrix}
\end{equation}

\noindent where $F_{INS}^n$ is the system matrix and its contents are shared in detail in \cite{ref29}. In the studies, this continuous time state space model is implemented by transforming it into a discrete time state space model.

\end{document}